\documentclass[11pt]{article}

\usepackage[final]{acl}

\usepackage{times}
\usepackage{latexsym}
\usepackage{amsmath}
\usepackage{amsfonts}
\usepackage{amssymb}
\usepackage{pifont}
\usepackage{booktabs}
\usepackage{tabularx}
\usepackage{multirow}

\usepackage{pgfplots}
\pgfplotsset{compat=1.17}

\usepackage{tikz}
\usepackage{xcolor}
\usepackage{array}

\newcommand{\idxfig}[2]{\textsc{#1}{\scriptsize #2}}  % For figures

\usetikzlibrary{shapes.geometric, arrows, positioning, fit, backgrounds}

\usepackage[T1]{fontenc}

\usepackage[utf8]{inputenc}

\usepackage{microtype}

\usepackage{inconsolata}

\usepackage{graphicx}

\usepackage{enumitem}

\title{DiscoSign: Discourse-Aware Text to Sign Language Gloss Translation}

\author{
  Vasileios Baltatzis \thanks{Equal contribution.}  \\
  Apple \\
  \And
  Mert Inan \footnotemark[1]
  \thanks{Work done entirely at Apple.}\\
  Northeastern University \\
  \And
  Connor Gillis \\
  Apple \\
  \AND
  Raja Kushalnagar \footnotemark[2] \\
  Gallaudet University \\
  \And
  Lorna Quandt \footnotemark[2] \\
  Gallaudet University \\
\And
  Leah Findlater \\
  Apple \\
  \And
  Colin Lea \\
  Apple \\
  }

\begin{document}
\maketitle

\begin{abstract}
Sign language processing systems have traditionally operated at the sentence level, ignoring critical discourse phenomena fundamental to sign language comprehension. We introduce DiscoSign, a computational approach for discourse-aware text to sign language gloss translation grounded in linguistic research. We address three key phenomena within our modular Large Language Model (LLM)-based translation framework: (i) spatial coreference resolution, where entities maintain consistent spatial locations throughout discourse; (ii) Question-Answer Clauses (QACs), pseudocleft structures serving specific discourse functions; and (iii) concept-gloss consistency, ensuring stable mappings between English concepts and American Sign Language (ASL) signs. 
Traditional translation metrics fail to capture discourse-level quality, so we introduce a suite of novel evaluation metrics designed to assess each dimension of discourse coherence addressed by our framework.
Experiments on sentence-level and discourse-level datasets show that our approach for discourse-aware processing significantly improves spatial consistency and entity tracking relative to sentence-only translation, while maintaining competitive single-sentence gloss translation quality. 
Our work establishes the first systematic framework for discourse-level text to sign language gloss translation with corresponding evaluation methodology.
\end{abstract}

% !TEX root = ../main.tex

\section{Introduction}
\label{sec:introduction}

\begin{figure}[t!]
\fontsize{7.5pt}{9pt}
\centering
\begin{tikzpicture}[
node distance=0.15cm,
box/.style={rectangle, draw, thick, text width=\columnwidth-0.4cm, align=left, inner sep=3pt},
englishbox/.style={box, draw=gray, fill=gray!10},
problembox/.style={box, draw=red!70, fill=red!5},
solutionbox/.style={box, draw=green!70, fill=green!5},
legendbox/.style={box, draw=gray, fill=gray!5, align=center, font=\scriptsize},
]
\setlength{\fboxsep}{1pt}
\node[englishbox] (english) {
\textbf{English:}
\textbf{S1:} ``Alice needed a car.''
\textbf{S2:} ``She asked Bob for his car.''
\textbf{S3:} ``He agreed as he would take the bus.''
};

\node[problembox, below=0.2cm of english] (problem) {
\textbf{Sentence-Level}\\[0.1cm]
\textbf{S1:} \textsc{alice}
\raisebox{-0.2cm}{{\setlength{\fboxsep}{1pt}\fcolorbox{gray}{white}{\includegraphics[height=0.7cm]{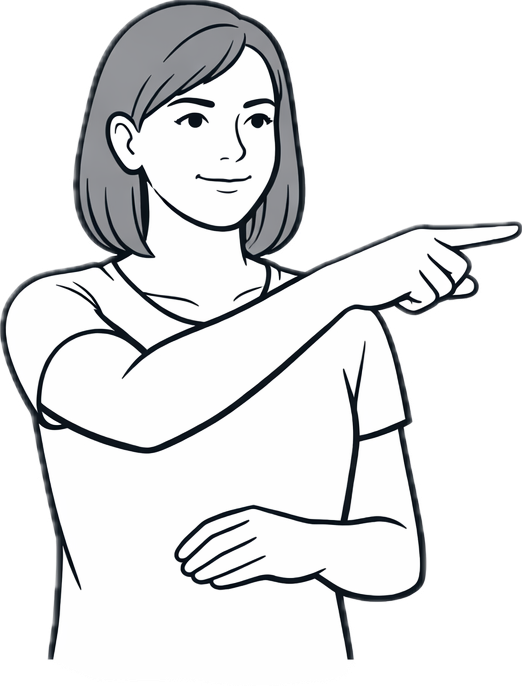}}}}
\fcolorbox{gray}{white}{\idxfig{ix-}{3p:i}}
\textsc{need}
\fcolorbox{gray}{gray!20}{\textsc{car}}\\[0.1cm]
\textbf{S2:} 
\raisebox{-0.2cm}{{\setlength{\fboxsep}{1pt}\fcolorbox{green}{white}{\includegraphics[height=0.7cm]{img/signer_i.png}}}}
\fcolorbox{green}{white}{\idxfig{ix-}{3p:i}}
\textsc{ask}
\raisebox{-0.2cm}{{\setlength{\fboxsep}{1pt}\fcolorbox{gray}{white}{\includegraphics[height=0.7cm]{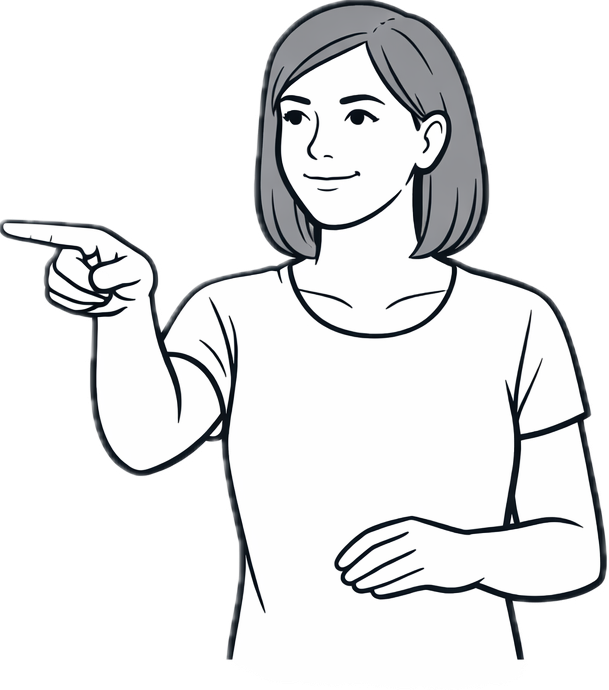}}}}
\fcolorbox{gray}{white}{\textsc{bob} \idxfig{ix-}{3p:j}}
\textsc{poss-}j \fcolorbox{red}{red!20}{\textsc{vehicle}}\\[0.1cm]
\textbf{S3:}
\raisebox{-0.2cm}{{\setlength{\fboxsep}{1pt}\fcolorbox{red}{white}{\includegraphics[height=0.7cm]{img/signer_i.png}}}}
\fcolorbox{red}{white}{\idxfig{ix-}{3p:i}}
\textsc{agree}
\raisebox{-0.2cm}{{\setlength{\fboxsep}{1pt}\fcolorbox{red}{white}{\includegraphics[height=0.7cm]{img/signer_i.png}}}}
\fcolorbox{red}{white}{\idxfig{ix-}{3p:i}}
\textsc{take bus}\\[0.1cm]
{\scriptsize\color{red}\ding{55} \idxfig{ix-}{3p:i} for Bob in S3; \textsc{car}$\rightarrow$\textsc{vehicle}; Missing causal QAC}
};

\node[solutionbox, below=0.2cm of problem] (solution) {
\textbf{Proposed}\\[0.1cm]
\textbf{S1:} \textsc{alice}
\raisebox{-0.2cm}{{\setlength{\fboxsep}{1pt}\fcolorbox{gray}{white}{\includegraphics[height=0.7cm]{img/signer_i.png}}}}
\fcolorbox{gray}{white}{\idxfig{ix-}{3p:i}}
\textsc{need}
\fcolorbox{gray}{gray!20}{\textsc{car}}\\[0.1cm]
\textbf{S2:} 
\raisebox{-0.2cm}{{\setlength{\fboxsep}{1pt}\fcolorbox{green}{white}{\includegraphics[height=0.7cm]{img/signer_i.png}}}}
\fcolorbox{green}{white}{\idxfig{ix-}{3p:i}}
\textsc{ask}
\raisebox{-0.2cm}{{\setlength{\fboxsep}{1pt}\fcolorbox{gray}{white}{\includegraphics[height=0.7cm]{img/signer_j.png}}}}
\fcolorbox{gray}{white}{\textsc{bob} \idxfig{ix-}{3p:j}}
\textsc{poss-}j
\fcolorbox{green}{green!20}{\textsc{car}}\\[0.1cm]
\textbf{S3:}
\raisebox{-0.2cm}{{\setlength{\fboxsep}{1pt}\fcolorbox{green}{white}{\includegraphics[height=0.7cm]{img/signer_j.png}}}}
\fcolorbox{green}{white}{\idxfig{ix-}{3p:j}}
\textsc{agree} \fcolorbox{green}{green!20}{\textsc{why?}}
\raisebox{-0.2cm}{{\setlength{\fboxsep}{1pt}\fcolorbox{green}{white}{\includegraphics[height=0.7cm]{img/signer_j.png}}}}
\fcolorbox{green}{white}{\idxfig{ix-}{3p:j}}
\textsc{take bus}\\[0.1cm]
{\scriptsize\color{green!50!black}\ding{51} Alice=i, Bob=j; \textsc{car} consistent; Causal QAC used}
};

\node[legendbox, below=0.2cm of solution] (legend) {
\raisebox{-0.15cm}{\includegraphics[height=0.5cm]{img/signer_i.png}} = \idxfig{ix-}{3p:i} \quad
\raisebox{-0.15cm}{\includegraphics[height=0.5cm]{img/signer_j.png}} = \idxfig{ix-}{3p:j} \quad
\fcolorbox{gray}{white}{\scriptsize idx} / \fcolorbox{gray}{gray!20}{\scriptsize gloss} = first definition \\ \quad
\fcolorbox{green}{white}{\scriptsize idx} / \fcolorbox{green}{green!20}{\scriptsize gloss} = correct \quad
\fcolorbox{red}{white}{\scriptsize idx} / \fcolorbox{red}{red!20}{\scriptsize gloss} = error
};

\end{tikzpicture}
\caption{Sentence-level translation violates spatial coreference and concept-gloss consistency, and misses a context where a QAC is appropriate. Our framework maintains discourse coherence through consistent spatial indices, stable concept-gloss mappings, and appropriate QAC usage.}
\label{fig:framework}
\end{figure}

Current approaches in text-to-sign language generation and translation systems predominantly operate at the sentence level. Each sentence is translated independently, without mechanisms to track how entities, concepts, and rhetorical choices should remain consistent across a multi-sentence text. As a result, these systems fail to capture discourse-level phenomena that are essential for coherent communication, which can have direct consequences for practical sign translation systems.

Most such systems rely on text-to-gloss translation as an intermediate step. Glosses are written labels denoting individual signs (e.g., CAT, BIRD), providing a representation that can be mapped to sign dictionaries for downstream sign production. The available gloss-sign dictionary defines the vocabulary constraints for the text-to-gloss task and generated glosses must correspond to entries in the dictionary. Discourse-level errors introduced at the gloss stage propagate through the entire pipeline.

Signed languages, such as American Sign Language (ASL), employ sophisticated grammatical structures including spatial indexing and rhetorical questions, which have discourse-level consistency requirements. \textit{Spatial coreference} relies on a complex spatial reference system where entities are assigned specific spatial coordinates (e.g., ASL glosses of: \textit{IX-3p:i}, \textit{IX-3p:j} referring to third person objects/subjects)\footnote{We use index notation like \textit{IX-3p:i}, where \textit{IX} here refers to index pointing, \textit{3p} refers to third-person, and \textit{i} refers to the i-th arbitrary location in space.} that must remain consistent throughout discourse \cite{winston1991spatial}. A signer may establish several referents in a scene and later reference them by pointing to the physical locations that were previously established. Sentence-level systems have no mechanism to track these assignments as a referent established as \textit{IX-3p:i} in one sentence may be arbitrarily reassigned to a different location in the next.  \textit{Question-Answer Clauses (QACs)}, or pseudoclefts, are sentence structures used to emphasize information, starting with a rhetorical question word like "what" or "where" and then providing the answer. The English declarative "I want coffee" could be phrased as "What I want is coffee," and in ASL gloss as "I WANT WHAT? COFFEE." QAC usage varies depending on discourse context and information structure; without access to this context, systems cannot identify when such restructuring is appropriate. \textit{Concept-gloss consistency} requires that the same concept map to the same gloss throughout discourse to maintain cohesion. For example, the English word “\textit{computer}” appearing five times should be rendered with the same ASL sign throughout, rather than alternating between \textit{COMPUTER}, \textit{MACHINE}, and \textit{TECHNOLOGY}. Sentence-level systems, lacking this context, may render the same concept inconsistently across sentences.

In this paper, we introduce, DiscoSign, the first framework of discourse-aware text-to-gloss translation. This work makes four primary contributions to computational sign language processing:

\begin{enumerate}[noitemsep]
    \item \textbf{Discourse-aware text-to-gloss translation framework}: We present the first systematic approach to sign language gloss translation from text that maintains cross-sentence coherence through explicit state registries and programmatic constraint enforcement for spatial coreference resolution, pseudocleft usage, and concept-gloss consistency.

    \item \textbf{ASL-centric instruction design methodology}: We develop a systematic approach to embedding specialized ASL linguistic expertise into Large Language Model (LLM) prompts, demonstrating how complex discourse phenomena can be addressed through careful instruction design.

    \item \textbf{Comprehensive evaluation framework}: We establish novel evaluation metrics for discourse-level text-to-gloss translation including spatial indexing consistency measurement, pseudocleft appropriateness assessment, and concept-gloss mapping consistency evaluation through cross-sentence reasoning tasks.
    
    \item \textbf{Systematic instruction component analysis}: We provide detailed ablation studies demonstrating the individual and combined contributions of different instruction components, establishing the relative importance of spatial indexing consistency, and discourse phenomena for translation quality, and we show that these gains hold for both a proprietary and an open-weight LLM backbone.
\end{enumerate}

% !TEX root = ../main.tex
\section{Related Work}
\paragraph{Sign Language Linguistics and Processing}

Sign languages exhibit rich linguistic structures that present unique challenges for computational modeling \cite{yin-etal-2021-including}. Prior work has addressed various linguistic phenomena including coreference resolution with spatial agreement \cite{yin-etal-2021-signed} and non-manual markers such as facial expressions and prosodic intensification \cite{10.1145/3706598.3713855, viegas-etal-2023-including, inan-etal-2022-modeling}, alongside system-level and production-side considerations for signing avatars \cite{hait-campbell_2025, Baltatzis_2024_CVPR}. The challenges facing sign language NLP are compounded by broader issues affecting low-resource languages \cite{court-elsner-2024-shortcomings, zhong2025opportunitieschallengeslargelanguage} and demographic biases in model performance \cite{atwell-etal-2024-studying}. Discourse-relevant phenomena were also identified in earlier computational work on sign language translation: \citet{stein-etal-2007-hand} showed that the spatial position of deictic signs carries referential information that helps disambiguate pronouns when translating from a signed language into English, and \citet{lugaresi2013translating} analyzed how Italian discourse connectives are realized in Italian Sign Language, observing that inter-sentential connectives are frequently dropped. Both establish that reference and discourse relations behave differently in signed languages and must be handled explicitly. Our work takes up the complementary computational problem of \textit{tracking and enforcing} such discourse coherence across multi-sentence texts, through shared state that persists between sentences and metrics that measure whether coherence is maintained. The pragmatics of signed discourse presents particularly rich phenomena: \citet{caponigro-davidson-2011} analyzed Question-Answer Clauses (QACs) as topic-comment structures governed by Question Under Discussion theory, while \citet{wilbur1994foregrounding} identified foregrounding structures where Q-constituents convey background information and A-constituents introduce new information. Advances in multilingual discourse prediction \cite{eichin-etal-2025-probing} suggest promising directions for modeling such phenomena. Building on this rich linguistic background, we develop the first discourse-aware sign language generation framework.

\paragraph{LLMs for Sign Language Translation}

The integration of large language models into sign language processing has emerged as a promising research direction. Recent work has demonstrated that LLMs possess inherent capabilities for sign language translation \cite{Gong_2024_CVPR}, with approaches ranging from gloss-free translation via direct visual-to-text mapping \cite{wong2024sign2gptleveraginglargelanguage, hwang2025efficientsignlanguagetranslation} to vocabulary sharing strategies for cross-modal knowledge transfer \cite{lee_kim_hwang_kim_park_2023}. Training strategies have focused on semantically aware label smoothing \cite{fayyazsanavi2024gloss2textsignlanguagegloss} and techniques for low-resource scenarios \cite{bulla2025leveraginglargelanguagemodels}. More recently, \citet{inan-etal-2025-signalignlm} introduced SignAlignLM, the first framework to natively integrate multimodal sign language processing into LLMs, in addition to practical deployment in large-scale dialogue systems \cite{inan-etal-2024-generating}. While these advances have improved translation accuracy, existing approaches predominantly focus on sentence-level translation and neglect discourse-level pragmatic phenomena essential for natural signed communication.

\paragraph{Evaluation of Sign Language Systems}

Evaluation methods for sign language systems have historically relied on metrics borrowed from machine translation, which fail to capture visual-spatial and discourse-level properties of signed languages \cite{yin-etal-2021-including}. Standard metrics such as BLEU \cite{papineni2002bleu} or ROUGE \cite{lin2004rouge} are insufficient for assessing sign language translation quality, as they focus on lexical overlap and penalize translations that exhibit various nuances of sign languages (e.g. flexible word order, omission of function words, and spatial-visual elements) \cite{muller2023considerations, Lea_2026_CVPR}. Metrics such as chrF \cite{popovic2015chrf} which compares character n-grams and is inherently more robust to word order differences, or COMET \cite{rei2020comet}, which captures semantic similarity and aligns with human judgment, are better suited. 
\citet{imai2025silverscoresemanticallyawareembeddingssign} introduced SilverScore, a semantically aware embedding-based metric demonstrating robustness to semantic and prosodic variations. However, existing metrics cannot evaluate pragmatic appropriateness or discourse coherence, failing to distinguish semantically correct but pragmatically awkward translations, such as using declarative word order where QAC structures are preferred \cite{caponigro-davidson-2011, wilbur1994foregrounding}. Our proposed evaluation suite (Section~\ref{subsec:framework_validation}) addresses these gaps through novel metrics for discourse-level phenomena specific to signed languages.

% !TEX root = ../main.tex

\section{Discourse-Aware Sign Language Gloss Translation Framework}
\label{sec:framework}

We introduce DiscoSign, the first computational framework for discourse-aware text to sign language gloss translation. Section~\ref{subsec:linguistic_foundations} presents the linguistic foundations underlying our approach, detailing the three discourse phenomena that current systems fail to address: spatial coreference, question-answer clauses, and concept-gloss consistency. Section~\ref{subsec:framework_architecture} then describes our modular framework, which operationalizes each phenomenon through a dedicated module with explicit state representations and constraint enforcement.

\subsection{Linguistic Foundations}
\label{subsec:linguistic_foundations}

This section presents our framework's linguistic foundations, formalizing three phenomena central to sign language discourse coherence: spatial coreference, question-answer clauses, and concept-gloss consistency. We illustrate these using the example in Figure~\ref{fig:framework}.

\paragraph{Spatial Coreference Resolution} 
Sign language discourse requires a unified spatial reference system that maintains both entity tracking and spatial consistency \cite{yin-etal-2021-signed}. Unlike spoken languages that rely on morphological agreement, sign languages use spatial indexing where entities are assigned consistent spatial locations (\textit{IX-3p:i}, \textit{IX-3p:j}) and directional relationships (\textit{i:GIVE:j}, \textit{1p:MEET:i}) throughout discourse. This system must coordinate entity coreference resolution with spatial grammar, ensuring that pronouns, definite descriptions, and repeated mentions resolve to correct antecedents through appropriate spatial index coordination across sentence boundaries. In Figure \ref{fig:framework}, \textit{Alice} is established at \textit{IX-3p:i} and \textit{Bob} at \textit{IX-3p:j}; subsequent pronouns must reference these locations consistently, and the possessive "\textit{his}" must be rendered as \textit{POSS-j} to maintain the link to \textit{Bob}.

\paragraph{Question-Answer Clauses (QACs)} 
QACs are topic-comment structures that are widely-occurring grammatical components of many sign languages \cite{wilbur1994foregrounding, caponigro-davidson-2011}. 
In American Sign Language, this topic-comment structure is specialized into QACs, which consist of a preamble question that is directly answered by a following declarative statement \cite{caponigro-davidson-2011}. From a syntactic perspective, QACs are argued to be pseudocleft structures with silent copula \cite{wilbur1994foregrounding}.

QACs serve specific discourse functions and must align with contextual requirements for natural sign language discourse. They are appropriate in contexts such as causal explanations, topic introductions, or emphasis, but should not appear at discourse beginnings or as direct responses to genuine questions. Awkward QAC usage can disrupt discourse flow and reduce comprehension \cite{wilbur1994foregrounding}. Identifying when QAC restructuring is appropriate therefore requires access to discourse context that sentence-level systems lack. In Figure \ref{fig:framework}, the English clause "as he would take the bus" expresses \textit{Bob's} reason for agreeing; the discourse-aware translation restructures this as a QAC that topicalizes the causal relationship: \textit{"IX-3p:j AGREE WHY? IX-3p:j TAKE BUS"}.

\paragraph{Concept-Gloss Consistency} Discourse coherence requires stable concept-to-sign mappings to reduce cognitive load and maintain semantic clarity \cite{halliday1976cohesion, frederiksen2019referential}. The same English concepts should consistently translate to the same ASL glosses throughout discourse while respecting vocabulary constraints. In Figure \ref{fig:framework} the English passage uses both \textit{car} and \textit{vehicle} to refer to the same object. A discourse-aware system recognizes this coreference and renders both as \textit{CAR}; a sentence-level system might produce \textit{CAR} and \textit{VEHICLE}, creating ambiguity about whether two objects are being discussed.

These phenomena interact throughout discourse: spatial coreference and concept-gloss consistency both require entity tracking, while QAC usage depends on information spanning multiple clauses. The following sections describe how our framework addresses each through dedicated modules.

\subsection{DiscoSign Modules}
\label{subsec:framework_architecture}

\begin{figure}[t]
\centering
\begin{tikzpicture}[
  font=\small,
  box/.style={draw, rounded corners=2pt, align=center, inner sep=4pt,
              minimum height=7mm, text width=36mm},
  reg/.style={draw, rounded corners=2pt, align=center, inner sep=4pt,
              fill=black!5, text width=22mm},
  arr/.style={-latex, thick},
]
\node[box] (sent) {Sentence $s_t$};
\node[box, below=5mm of sent] (prompt)
  {Constrained prompt\\[1pt] {\scriptsize registries as hard constraints}};
\node[box, below=5mm of prompt] (llm)
  {LLM call $\rightarrow$ JSON\\[1pt] {\scriptsize gloss $+$ structured metadata}};
\node[box, below=5mm of llm] (post)
  {Deterministic verification\\[1pt] {\scriptsize correct violations in place}};
\node[box, below=5mm of post] (out) {Gloss $\hat{g}_t$};
\node[reg, right=7mm of prompt] (reg)
  {Registries\\ $\mathcal{S}$, $\mathcal{L}$, $\mathcal{Q}$\\[1pt]
   {\scriptsize from $s_1 \ldots s_{t-1}$}};

\draw[arr] (sent) -- (prompt);
\draw[arr] (prompt) -- (llm);
\draw[arr] (llm) -- (post);
\draw[arr] (post) -- (out);
\draw[arr] (reg) -- (prompt);
\draw[arr] (post.east) -| node[pos=0.3, above, font=\scriptsize] {update for $s_{t+1}$} (reg.south);
\end{tikzpicture}
\caption{The DiscoSign pipeline. Each sentence is translated in a single LLM call whose prompt states the accumulated discourse registries as hard constraints. A deterministic post-processor corrects any remaining violations, and the verified metadata updates the registries for the next sentence.}
\label{fig:pipeline}
\end{figure}
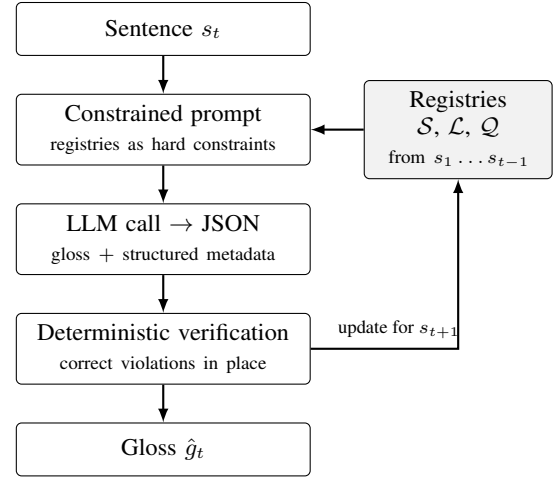

We operationalize these linguistic phenomena through three interconnected modules, each with an explicit state representation and a set of constraints. Before giving the formal definitions, we describe the end-to-end pipeline, shown in Figure~\ref{fig:pipeline}.

The framework translates a text one sentence at a time, carrying discourse state forward in three registries: $\mathcal{S}$ for spatial assignments, $\mathcal{L}$ for concept-gloss mappings, and $\mathcal{Q}$ for QAC decisions. For each sentence $s_t$, the framework (i)~retrieves the registries accumulated over sentences $s_1 \ldots s_{t-1}$; (ii)~constructs a single prompt containing $s_t$, a context window of previous sentence--translation pairs, and the registry contents stated as hard constraints; (iii)~receives a JSON response with the gloss translation and structured metadata (spatial mappings, concept mappings, and a QAC flag); (iv)~runs a deterministic post-processor that checks the output against the registries and corrects violations by string replacement, without re-querying the model; and (v)~updates the registries from the verified metadata, so that they constrain $s_{t+1}$. The modules are therefore not separate model calls: each is a structured prompt section together with its registry and verification logic, all within a single LLM call per sentence.

The rest of this section formalizes each module's state and constraints. Prompt templates are given in Appendix~\ref{app:prompts} and the verification procedure in Appendix~\ref{app:verification}.

\paragraph{Spatial Coreference Module (SCM)}
\label{subsubsec:scm}
Given a new sentence, the SCM identifies entity mentions, resolves coreference with previously established entities, and either retrieves existing spatial assignments or assigns new indices for novel entities. The module outputs spatially-indexed glosses (e.g., \textit{IX-3p:i}, \textit{POSS-j}) that maintain referential consistency throughout the discourse.
We formalize this by defining the module's state as tracking spatial assignments and their corresponding entities:
\begin{equation}
\mathcal{S} = \{(e, i, t) : e \in \mathcal{E}, i \in \mathcal{I}, t \in \mathcal{T}\}
\end{equation}

where $e$ is an entity, $i$ is a spatial index, and $t$ is the sentence where the assignment was established.

The SCM enforces two primary constraints:
\begin{itemize}[nolistsep]
    \item \textit{Spatial Consistency}: For any entity $e$, if $(e, i, t) \in \mathcal{S}$, then all subsequent references to $e$ in sentences $t' > t$ use spatial index $i$.
    
    \item \textit{Coreference Correspondence}: For any spatial reference claiming entity $e$ uses index $i$, there must exist a coreference chain $\mathcal{C}_e$ in the English text that includes the claimed mention.
\end{itemize}

For directional verbs $i_{source}:VERB:i_{target}$, both source and target indices must satisfy these constraints independently.

\paragraph{Question-Answer Clause Module (QACM)}
\label{subsubsec:qacm}
Given a new sentence and its discourse context, the QACM analyzes whether the sentence contains information suitable for topicalization, such as causal explanations, contrastive elements, or focal points. If appropriate, the module selects the corresponding QAC type and restructures the output accordingly, ensuring that QAC usage patterns remain consistent with ASL discourse conventions throughout the text.
We formalize this by defining the module's state as a registry of QAC decisions across the discourse:
\begin{equation}
\mathcal{Q} = \{(t, q) : t \in \mathcal{T}, q \in \{0,1\}\}
\end{equation}
where $t$ is the sentence index and $q$ indicates whether a QAC structure was used. The QACM enforces two constraints:
\begin{itemize}[noitemsep]
    \item \textit{Discourse Position}: QACs should not appear at discourse beginnings ($t{=}0$) or as responses to genuine questions in the source text.
    
    \item \textit{Contextual Appropriateness}: QAC insertion depends on information structure spanning multiple clauses; the module evaluates preceding discourse to determine whether topicalization is warranted.
\end{itemize}

\paragraph{Concept-Gloss Consistency Module (CGCM)}
\label{subsubsec:cgcm}
Given a new sentence, the CGCM identifies content words and checks whether they correspond to concepts already established in prior context. For known concepts, the module retrieves and applies the existing gloss mapping. For novel concepts, it selects an appropriate gloss from the available vocabulary and registers the new mapping. This process ensures that synonyms or paraphrases in the English source (e.g., ``car'' and ``vehicle'') resolve to a single consistent gloss throughout discourse.
We formalize the module's state as a concept registry $\mathcal{L}$ that tracks established mappings:
\begin{equation}
\mathcal{L} = \{(c, g, t) : c \in \mathcal{C}, g \in \mathcal{V}, t \in \mathcal{T}\}
\end{equation}

where $c$ is an English concept, $g$ is an ASL gloss from vocabulary $\mathcal{V}$, and $t$ is the sentence where the mapping was established. 
The CGCM enforces:
\begin{itemize}[noitemsep]
    \item \textit{Mapping Consistency}: For any concept $c$, if $(c, g, t) \in \mathcal{L}$, then all subsequent translations of $c$ in sentences $t' > t$ must use gloss $g$.
    
    \item \textit{Vocabulary Compliance}: All glosses must satisfy $g \in \mathcal{V}$, ensuring mappings respect the constraints imposed by available dictionaries.
\end{itemize}

\paragraph{Integrated Processing}
\label{subsec:integrated_processing}
The framework coordinates all modules through a unified state $ \mathcal{U} = (\mathcal{S}, \mathcal{L}, \mathcal{Q})$.
For each input sentence, the three modules operate in coordination. Each module's decisions are informed by and constrained by the others. For instance, a QAC restructuring must preserve the spatial indices established by the SCM. 

The unified state $\mathcal{U}$ is updated after each sentence: the structured output (spatial mappings, concept mappings, QAC usage) is extracted and used to update the registries for subsequent sentences.

\paragraph{Constraint Enforcement}
\label{subsec:constraint_enforcement}
Module constraints are enforced programmatically rather than relying solely on LLM instruction-following. Before each LLM call, the prompt is constructed as a deterministic function of the current registry state and sentence position. After each call, a verification step checks the output against the registries: spatial index and concept-gloss violations are corrected via string replacement throughout the translation. All enforcement is deterministic and requires no re-querying (Appendix~\ref{app:verification}).

\section{Evaluation Suite}
\label{subsec:framework_validation}
We introduce novel metrics aligned with each module of our framework. Each metric evaluates a specific aspect of discourse coherence that traditional translation metrics cannot capture, providing the first systematic evaluation of cross-sentence phenomena in text to sign language gloss translation.

\subsection{Spatial Coreference Accuracy (SCA)}
\label{subsubsec:sca_metric}
SCA evaluates the SCM by measuring how accurately spatial indexing in ASL translations preserves coreference relationships from the source English discourse. Intuitively, if \textit{Alice} is initially assigned to location \textit{IX-3p:i}, all subsequent references to \textit{Alice} must use the same location. We evaluate spatial references excluding first definitions, where entities are initially assigned to spatial locations. For directional verbs (\textit{i:VERB:j}), both source and target indices are evaluated independently. For each spatial reference $r$ claiming entity $e$ at index $i$ in sentence $t$, we perform two checks.

The \textit{spatial consistency check} verifies that entity $e$ has not been previously assigned a different spatial index:
\begin{equation}
\begin{aligned}
\text{spatial\_consistent}(r) = &\nexists (e, i', t') \in \mathcal{S} : \\
&i' \neq i \land t' < t
\end{aligned}
\end{equation}

The \textit{coreference correspondence check} verifies that the claimed entity $e$ has a mention in the aligned English sentence according to the coreference chain $\mathcal{C}_e$:
\begin{equation}
\begin{aligned}
\text{coreference\_valid}(r) = &\exists m \in \mathcal{C}_e : \\
&m \text{ appears in sentence } t
\end{aligned}
\end{equation}

SCA is calculated as the proportion of spatial references passing both checks:
\begin{equation}
\begin{aligned}
\text{SCA} = \frac{1}{|R|} \sum_{r \in R} [\text{spatial\_consistent}(r) \land \\ \text{coreference\_valid}(r)]
\end{aligned}
\end{equation}
where $R$ represents all spatial references in the ASL translation. SCA addresses the core challenge of ASL discourse translation: maintaining consistent spatial-entity mappings across sentences while preserving referential relationships. This metric captures phenomena that traditional word-level metrics cannot assess, providing the first systematic evaluation of ASL discourse coherence.

\subsection{Question-Answer Clause Appropriateness
\texorpdfstring{(QAC$_{\text{Ap}}$)}{QACAP}}
\label{subsubsec:qacap_metric}

QAC$_{\text{Ap}}$ evaluates the QACM by measuring whether QACs appear in linguistically suitable contexts. For instance, a QAC is appropriate when translating ``He agreed because he would take the bus'' (causal relationship) but not when directly answering a genuine question like ``Why did he agree?'' in the source text. For each sentence $t \in \mathcal{T}$ where a QAC is used, we evaluate:
\begin{equation}
\begin{aligned}
\text{appropriate}(t) = &\mathbb{I}[\neg\text{answersQuestion}(t-1)] \land \\ & \mathbb{I}[\text{hasContext}(t)]
\end{aligned}
\end{equation}

where $\mathbb{I}[\cdot]$ is the indicator function that returns 1 if the condition is true and 0 otherwise, $\text{hasContext}(t)$ indicates the presence of causal, topical, or emphasis discourse markers, and the question-answering check applies only when $t > 1$. The overall QAC appropriateness is calculated as:
\begin{equation}
\text{QAC}_{\text{Ap}} = \frac{\sum_{t \in \mathcal{T}_{QAC}} \text{appropriate}(t)}{|\mathcal{T}_{QAC}|}
\end{equation}

where $\mathcal{T}_{QAC} \subseteq \mathcal{T}$ represents the set of sentences where QACs were used. This metric focuses on precision rather than recall, evaluating whether QAC usage decisions were contextually appropriate rather than penalizing missed opportunities, as QACs are optional discourse enhancement tools.

\subsection{Concept-Gloss Consistency (CGC)}
\label{subsubsec:cgc_metric}

CGC evaluates the CGCM by measuring whether English concepts receive consistent ASL gloss translations throughout discourse. Inconsistent mappings, such as rendering ``car'' as \textit{CAR} in one sentence and \textit{VEHICLE} in another, can disrupt a signers's ability to track entities across discourse. For each concept $c$ appearing in sentence $t$ as gloss $g$, we verify consistency with prior occurrences:
\begin{equation}
\begin{aligned}
\text{concept\_consistent}(c, g, t) = &\nexists (c, g', t') \in \mathcal{L} : \\
&g' \neq g \land t' < t
\end{aligned}
\end{equation}

CGC is calculated as the proportion of repeated concepts that maintain consistent mappings:
\begin{equation}
\text{CGC} = \frac{\sum_{c \in \mathcal{C}_{repeated}} \text{concept\_consistent}(c)}{|\mathcal{C}_{repeated}|}
\end{equation}

where $\mathcal{C}_{repeated}$ represents concepts that appear in multiple sentences. CGC evaluates concept-gloss consistency across all discourse-level concept repetitions, ensuring that gloss choices support discourse coherence. This validation captures the fundamental requirement that identical concepts should receive identical gloss translations throughout discourse.

% !TEX root = ../main.tex
\section{Experiments and Findings}

\label{sec:experiments}

We conduct comprehensive experiments to validate DiscoSign across two complementary evaluation paradigms: sentence-level translation accuracy and discourse-level coherence assessment.

\subsection{Experimental Setup}
\label{subsec:experimental_setup}

\textbf{Datasets.} We evaluate on three datasets spanning sentence-level and discourse-level phenomena (Table~\ref{tab:datasets}). ASL STEM Wiki and a licensed dataset provide sentence-level evaluation with expert-annotated ground truth glosses. The licensed dataset pairs English sentences describing everyday activities with ASL glosses, and was collected and annotated by a team of Deaf signers fluent in both ASL and English. ASL STEM Wiki contains scientific terminology requiring substantial fingerspelling. Aesop's Fables provides discourse-level evaluation with multi-sentence narratives containing coreference chains, causal relationships, and topic transitions.

\begin{table}[t]
\centering
\small
\begin{tabularx}{\columnwidth}{@{}l>{\raggedright\arraybackslash}rccc@{}}
\textbf{Dataset} & \textbf{Examples} & \textbf{Sents./Ex.} & \textbf{Gold} & \textbf{Disc.} \\
\midrule
ASL STEM Wiki  & 500 & 1 & \checkmark & \ding{55} \\
Licensed Dataset & 2121 & 1 & \checkmark & \ding{55} \\
Aesop's Fables  & 284 & 5 & \ding{55} & \checkmark \\
\bottomrule
\end{tabularx}
\caption[]{Evaluation datasets. Examples: sentence pairs (ASL STEM Wiki \citep{yin2024asl}, Licensed Dataset) or texts (Aesop's Fables \citep{aesop1912gutenberg}). Sents./Ex.: average sentences per example. Gold: expert-annotated reference glosses available. Disc.: contains discourse-level phenomena.}
\label{tab:datasets}
\end{table}

\noindent \textbf{System Configurations.} We compare three system configurations: (1) \textbf{Sentence-level}: Gemini 2.5 Pro with access only to the current sentence and no discourse-specific instructions, (2) \textbf{Context-Aware}: the same model with access to discourse context (i.e. previous sentences) but no specialized discourse instructions, and (3) \textbf{Proposed Framework}: Our proposed framework with all three discourse modules (SCM, QACM, CGCM) implemented as structured prompt sections with explicit state registries and post-processing verification. All systems use the ASLLRP SignBank \cite{neidle2022asl} lexicon (2,859 glosses) as vocabulary constraints and deterministic sampling (temperature=0).

\noindent \textbf{Evaluation Details.} Every LLM call, both in the translation pipeline and in evaluation, uses Gemini 2.5 Pro unless stated otherwise; \S\ref{subsec:backbone} additionally reports the framework with an open-weight backbone. For back-translation evaluation, we prompt Gemini 2.5 Pro without discourse-specific instructions to translate generated glosses back to English, then compare against the original source. For discourse-level evaluation, the English coreference chains required by SCA were extracted with Gemini 2.5 Pro and then validated by a human expert. Similarly, the structured outputs generated during translation (spatial mappings, concept-gloss mappings) were verified by a human expert for faithfulness to the ASL translations. All prompts and configurations are provided in Appendix \ref{app:prompts}.

\subsection{Sentence-Level Translation Quality}
\label{subsec:sentence_level_experiments}

We first validate that our discourse-aware framework maintains competitive sentence-level translation quality on datasets where discourse phenomena are absent. Table~\ref{tab:sentence_results} presents results on both ASL STEM Wiki and the licensed dataset, providing both gloss-level and back-translation evaluation.
\begin{table}[t]
\resizebox{\columnwidth}{!}{
\centering
\begin{tabular}{llcccc}
 & & \multicolumn{2}{c}{\textbf{Gloss}} & \multicolumn{2}{c}{\textbf{Back-translation}} \\
 \cmidrule(lr){3-4} \cmidrule(lr){5-6}
\textbf{Dataset} & \textbf{Approach} & \textbf{chrF} & \textbf{COMET} & \textbf{chrF} & \textbf{COMET} \\
\midrule
\multirow{2}{*}{ASL STEM Wiki} & Sentence-level & 28.2 & \textbf{0.54} & 41.9 & 0.73 \\
 & Proposed & \textbf{30.1} & 0.51 & \textbf{54.8} & \textbf{0.81} \\
\midrule
\multirow{2}{*}{Licensed Dataset} & Sentence-level & \textbf{39.6} & \textbf{0.59} & 65.2 & 0.88 \\
 & Proposed & 39.1 & 0.57 & \textbf{67.2} & \textbf{0.90} \\
\bottomrule
\end{tabular}
}
\caption{Sentence-level translation results on ASL STEM Wiki and the licensed dataset.}
\label{tab:sentence_results}
\end{table}

\noindent \textbf{Discourse Integration Does Not Compromise Sentence-Level Quality.} Our proposed framework demonstrates competitive or superior performance across both datasets, confirming that discourse-aware capabilities do not degrade sentence-level translation. The lower absolute scores on ASL STEM Wiki reflect its challenging scientific vocabulary requiring extensive fingerspelling, compared to the licensed dataset's everyday language. Notably, the framework shows consistent improvements in back-translation quality (ASL STEM Wiki chrF: 54.8 vs 41.9, COMET: 0.81 vs 0.73; licensed dataset chrF: 67.2 vs 65.2, COMET: 0.90 vs 0.88), suggesting that the enhanced linguistic reasoning developed for discourse processing also benefits sentence-level semantic preservation.

\noindent \textbf{Gloss-Level and Back-Translation Metrics Capture Different Aspects.} 
Gloss-level evaluation yields lower scores than back-translation due to two factors: different datasets employ different gloss conventions, directly impacting string-matching metrics; and gloss metrics measure strict string similarity while back-translation tolerates surface-level variation. The higher back-translation scores indicate both approaches preserve meaning despite surface-level differences from gloss labels.

\subsection{Discourse-Level Translation Quality}
\label{subsec:discourse_level_experiments}
\label{subsec:backbone}

We evaluate discourse-aware translation on Aesop's Fables using our proposed metrics alongside traditional MT metrics. Table~\ref{tab:discourse_results} presents comparative results across all system configurations and both LLM backbones.

\noindent \textbf{Traditional Metrics Show Limited Sensitivity.}
All three Gemini-based approaches achieve comparable chrF (39.4–41.7) and COMET (0.76–0.78) scores despite dramatically different discourse coherence, confirming that traditional MT metrics cannot capture spatial consistency, coreference resolution, or pseudocleft structure requirements fundamental to sign language discourse comprehension.

\noindent \textbf{Explicit Discourse Modeling Outperforms Context Access Alone.} While providing context improves over no-context processing (SCA: 0.29 → 0.48; CGC: 0.59 → 0.68; QAC$_{\text{Ap}}$: 0.70 → 0.72), our structured framework achieves substantially higher performance across all discourse metrics (SCA: 0.84; CGC: 0.97; QAC$_{\text{Ap}}$: 0.76). This demonstrates that access to prior sentences is helpful but insufficient. Explicit modeling of spatial coreference resolution, pseudocleft structures, and concept-gloss consistency is necessary for discourse-aware translation. The dramatic improvements captured by SCA and CGC, coupled with the limited sensitivity of MT metrics, validate both our framework's effectiveness and the need for discourse-specific evaluation methodologies.

\begin{table}[t]
\resizebox{\columnwidth}{!}{
\centering
\begin{tabular}{llccccc}
\textbf{Backbone} & \textbf{Approach} & \textbf{SCA} & \textbf{CGC} & \textbf{QAC$_{\text{Ap}}$} & \textbf{chrF} & \textbf{COMET} \\
\midrule
\multirow{3}{*}{Gemini 2.5 Pro} & Sentence-Level & 0.29 & 0.59 & 0.70 & 40.9 & 0.77 \\
 & Context-Aware & 0.48 & 0.68 & 0.72 & 39.4 & 0.76 \\
 & Proposed & \textbf{0.84} & \textbf{0.97} & \textbf{0.76} & \textbf{41.7} & \textbf{0.78} \\
\midrule
\multirow{3}{*}{Qwen3.6-35B-A3B} & Sentence-Level & 0.12 & 0.73 & 0.73 & 40.1 & 0.73 \\
 & Context-Aware & 0.19 & 0.83 & 0.72 & \textbf{41.2} & 0.74 \\
 & Proposed & \textbf{0.61} & \textbf{0.86} & \textbf{0.75} & 40.3 & 0.74 \\
\bottomrule
\end{tabular}
}
\caption{Discourse-level translation results for both LLM backbones; the Qwen block is discussed in \S\ref{subsec:backbone}. With Gemini 2.5 Pro, all differences between Proposed and both baselines are statistically significant (p<0.05, paired bootstrap, 10k samples) except QAC$_{\text{Ap}}$ vs Context-aware (p=0.066) and COMET vs Sentence-level (p=0.345).}
\label{tab:discourse_results}
\end{table}

\noindent \textbf{Backbone Generalization.}
Our framework is LLM-agnostic by construction: any instruction-following model that can produce the specified JSON schema can serve as the backbone. We repeat the Aesop's Fables experiment with Qwen3.6-35B-A3B, an open-weight mixture-of-experts model (35B total, 3B active parameters), holding prompts and post-processing fixed. The ordering of the three conditions is preserved (Table~\ref{tab:discourse_results}): our framework again scores highest on the discourse metrics, improving SCA by $+$0.49 and CGC by $+$0.13 over the sentence-level baseline. Absolute scores are lower for the smaller model, most visibly on SCA. Traditional metrics again separate the conditions poorly, as under Qwen, the context-aware baseline attains the highest chrF despite the second-lowest SCA, reinforcing that our contribution lies in discourse coherence rather than surface-level quality.

\noindent \textbf{Qualitative Analysis.} Table~\ref{tab:qual_example} illustrates these differences concretely. The sentence-level baseline produces ``FOX'' in S1 but ``fs-W-E-A-S-E-L'' in S2 for the same entity, and assigns inconsistent spatial indices (\textit{IX-3p:j} in S1, \textit{IX-3p:i} in S2). Our framework maintains both concept-gloss consistency (``fs-W-E-A-S-E-L'' throughout) and spatial consistency (\textit{IX-3p:j} for \textit{Weasel} in both sentences).

\begin{table}[t]
\centering
\small
\setlength{\fboxsep}{1pt}
\begin{tabular}{p{0.9\columnwidth}}
\textbf{English:} \\
\textbf{S1:} A Bat fell to the ground and was caught by a Weasel, and was just going to be killed and eaten when it begged to be let go. \\
\textbf{S2:} The Weasel said he couldn't do that because he was an enemy of all birds on principle. \\
\midrule
\textbf{Sentence-level Baseline:} \\
\textbf{S1:} \textsc{bat} \idxfig{ix-}{3p:i} \textsc{fall ground} \idxfig{ix}{-loc:k}. \fcolorbox{red}{red!20}{\textsc{fox}} \fcolorbox{red}{white}{\idxfig{ix-}{3p:j}} \textsc{catch} \idxfig{ix-}{3p:i}. \idxfig{ix-}{3p:j} \textsc{almost kill eat} \idxfig{ix-}{3p:i}. \idxfig{ix-}{3p:i} \textsc{beg} \idxfig{ix-}{3p:j}. \textsc{for-for? free}. \\
\textbf{S2:} \fcolorbox{red}{red!20}{\textsc{fs-w-e-a-s-e-l}} \fcolorbox{red}{white}{\idxfig{ix-}{3p:i}} \textsc{say}  \textsc{cannot do that why} \idxfig{ix-}{3p:i} \textsc{always opposite+agent all bird} \\
\midrule
\textbf{Proposed:} \\
\textbf{S1:} \textsc{bat} \idxfig{ix-}{3p:i} \textsc{fall ground} \fcolorbox{green}{green!20}{\textsc{fs-w-e-a-s-e-l}} \fcolorbox{green}{white}{\idxfig{ix-}{3p:j}} \textsc{catch} \idxfig{ix-}{3p:i} \textsc{almost kill eat when} \idxfig{ix-}{3p:i} \textsc{beg free} \\
\textbf{S2:} \fcolorbox{green}{green!20}{\textsc{fs-w-e-a-s-e-l}} \fcolorbox{green}{white}{\idxfig{ix-}{3p:j}} \textsc{say cannot free} \idxfig{ix-}{3p:i} \textsc{why?} \idxfig{ix-}{3p:j} \textsc{opposite+agent all bird rule} \\
\end{tabular}
\caption{Qualitative examples. \fcolorbox{gray}{gray!20}{Filled boxes} = concepts/glosses; \fcolorbox{gray}{white}{unfilled boxes} = spatial indices. \textcolor{red}{Red} = error; \textcolor{green!50!black}{Green} = correct.}
\label{tab:qual_example}
\end{table}

\subsection{Ablation Studies}
\label{subsec:ablations}

\textbf{Each module plays a role.}  Table~\ref{tab:system_ablation} evaluates individual module contributions by selectively disabling each of SCM, QACM and CGCM. Each module primarily affects its corresponding metric: disabling SCM drops SCA from 0.81 to 0.71; disabling CGCM drops CGC from 0.97 to 0.64; with QAC$_{\text{Ap}}$ dropping from 0.76 to 0.71--0.72 when disabled. Notably, the presence of QACM is not as consequential on the translation metrics as its impact is mostly stylistic. The combination of all modules provides the most balanced performance across discourse-specific metrics, demonstrating that each module addresses distinct aspects of discourse coherence in ASL translation.
\begin{table}[t]
\resizebox{\columnwidth}{!}{
\centering
\begin{tabular}{ccc|ccccc}
\textbf{SCM} & \textbf{CGCM} & \textbf{QACM} & \textbf{SCA} & \textbf{CGC} & \textbf{QAC$_{\text{Ap}}$} & \textbf{chrF} & \textbf{COMET} \\
\midrule
\checkmark & \ding{55} & \ding{55} & 0.81 & 0.64 & 0.71 &41.8 & 0.78 \\
\ding{55} & \checkmark & \ding{55} & 0.71 & 0.95 & 0.72 &\textbf{42.5} & \textbf{0.79}\\
\checkmark & \checkmark & \ding{55} & \textbf{0.84} & 0.97  & 0.72&\textbf{42.5}  & 0.77 \\
\checkmark & \checkmark & \checkmark & \textbf{0.84} & \textbf{0.97} & \textbf{0.76} &41.7 & 0.78 \\
\bottomrule
\end{tabular}
}

\caption{Ablation study on each module's contribution.}
\label{tab:system_ablation}
\end{table}

\begin{figure}[t]
\centering
\small
\begin{tikzpicture}
\begin{axis}[
width=\columnwidth,
height=5.5cm,
xlabel={\small Context Window Size},
ylabel={\small Score},
xmin=-0.3, xmax=3.3,
ymin=0.0, ymax=1.02,
xtick={0,1,2,3},
xticklabels={0, 1, 2, Full},
ytick={0.1, 0.2, 0.3, 0.4, 0.5, 0.6, 0.7, 0.8, 0.9, 1.0},
legend pos=south east,
legend style={font=\footnotesize, cells={anchor=west}},
grid=major,
grid style={dashed, gray!30},
every axis plot/.append style={thick, mark size=2.5pt},
]

\addplot[
color=orange,
mark=square*,
] coordinates {
(0, 0.45)
(1, 0.57)
(2, 0.75)
(3, 0.84)
};

\addplot[
color=red,
mark=triangle*,
] coordinates {
(0, 0.78)
(1, 0.84)
(2, 0.87)
(3, 0.97)
};

\addplot[
color=blue,
mark=diamond*,
] coordinates {
(0, 0.72)
(1, 0.74)
(2, 0.76)
(3, 0.76)
};

\addplot[
color=orange,
mark=square,
only marks,
mark size=3pt,
mark options={fill=white, thick},
] coordinates {(0, 0.29)};

\addplot[
color=orange,
mark=square,
only marks,
mark size=3pt,
mark options={fill=white, thick},
] coordinates {(3, 0.47)};

\legend{SCA, CGC, QAC$_{\text{Ap}}$}

\node[font=\scriptsize, anchor=west] at (axis cs:0.1, 0.29) {Sentence-Level Baseline};
\node[font=\scriptsize, anchor=east] at (axis cs:2.9, 0.47) {Context-Aware Baseline};

\end{axis}
\end{tikzpicture}
\caption{Impact of context window size on discourse metrics. Unfilled markers show SCA for baselines without discourse-aware instructions.}
\label{fig:context_ablation}
\end{figure}
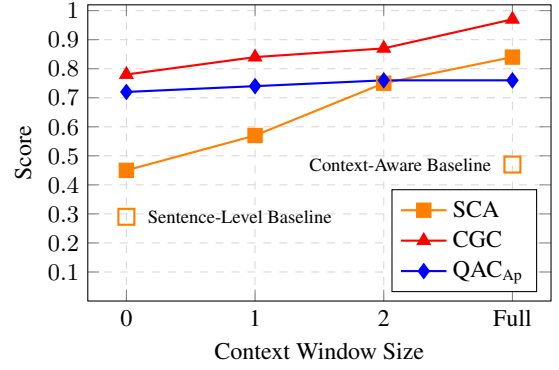

\noindent \textbf{Context Window Size.} Figure~\ref{fig:context_ablation} shows the impact of context. Increasing context progressively improves all discourse metrics, while traditional metrics remain unchanged (chrF: $\sim$42, COMET: $\sim$0.78 across all conditions). SCA shows the steepest improvement (0.40 → 0.84), CGC follows a similar pattern (0.78 → 0.97), and QAC$_{\text{Ap}}$ plateaus at two sentences (0.76). Comparing against baselines without discourse-aware instructions, our framework outperforms the sentence-level baseline even without context (SCA: 0.45 vs 0.29), and with just one prior sentence exceeds the context-aware baseline with full access (SCA: 0.57 vs 0.48). This demonstrates that structured modeling provides greater benefit than context quantity alone.

\subsection{Human Evaluation}
\label{subsec:human_eval}

To validate our proposed metrics, two ASL-fluent evaluators rated 32 stories (152 sentences) selected via stratified sampling across automated score ranges, drawn from both the proposed framework and sentence-level baseline. Sentences were rated on a 1--5 Likert scale for SCA, CGC, and QAC$_{\text{Ap}}$.

SCA shows the strongest validation, with significant story-level correlation for both evaluators individually ($\rho$\,=\,0.50, $p$\,=\,0.005 and $\rho$\,=\,0.42, $p$\,=\,0.022; combined $\rho$\,=\,0.52, $p$\,=\,0.003) and moderate inter-rater agreement (weighted $\kappa$\,=\,0.43, $\rho$\,=\,0.49). CGC shows significant correlation ($\rho$\,=\,0.48, $p$\,=\,0.007), with both evaluators individually trending in the same direction ($\rho$\,=\,0.48, $p$\,=\,0.007 and $\rho$\,=\,0.24, $p$\,=\,0.195). 
QAC$_{\text{Ap}}$ correlations are non-significant. However, QAC \textit{usage} agreement between the system's self-reported output and human judgments of QAC presence is strong for one evaluator
  ($\rho$\,=\,0.85, $p$\,$<$\,0.001) but moderate for the other ($\rho$\,=\,0.39, $p$\,=\,0.035). This suggests the non-significant appropriateness correlation reflects disagreement among raters about what constitutes a QAC rather than a failure of the automated metric (see Appendix~\ref{app:human_eval} for details).

% !TEX root = ../main.tex
\section{Conclusion}
\label{sec:conclusion}

This work establishes discourse-aware text to sign language gloss translation as a novel task and presents the first comprehensive framework to address it. We formalize three discourse phenomena: spatial coreference, question-answer clauses, and concept-gloss consistency, and develop specialized modules (SCM, QACM, CGCM) with corresponding evaluation metrics (SCA, QAC$_{\text{Ap}}$, CGC). Our experiments demonstrate that explicit discourse modeling substantially outperforms both sentence-level and context-only approaches, establishing a foundation for more linguistically authentic sign language translation systems.

% !TEX root = ../main.tex
\section{Limitations}

Our framework currently relies on large language models to implement each module, which introduces dependencies on LLM instruction-following capabilities and may limit reproducibility as model versions change. However, the modular framework we propose is agnostic to the specific implementation of each component, as our results with an open-weight backbone show (\S\ref{subsec:backbone}). SCM, QACM, and CGCM could alternatively be instantiated through rule-based systems, trained neural models, or human experts, depending on application requirements. Our evaluation focuses exclusively on English-to-ASL gloss translation. However, the linguistic phenomena we consider are not specific to ASL but are present across signed languages. Additionally, the Aesop's Fables dataset lacks ground truth ASL annotations, requiring us to rely on automatic metrics and back-translation for discourse-level evaluation; a reference-annotated discourse-level dataset would allow standard reference-based evaluation alongside our metrics, and we regard building one as a substantial annotation contribution in its own right. The QACM also enforces QAC decisions through deterministic trigger rules, which cannot represent the optionality of QAC usage in contexts where both a QAC and a non-QAC rendering are natural; we discuss learned alternatives in Appendix~\ref{app:human_eval}.

Finally, our framework addresses manual sign production through glosses but does not incorporate non-manual markers (NMMs) such as facial expressions and prosody, which carry critical grammatical and affective information in sign languages. We note, however, that the state each module maintains is closely related to the information a downstream visual generation system would need in order to produce NMMs: the QACM identifies question and answer constituents, the SCM assigns spatial indices to referents, and the CGCM tracks which referents are already established. Deriving NMM annotations from this registry state is a natural next step for connecting discourse-aware gloss translation to sign production.

\section{Ethical Considerations}
Sign language processing requires careful consideration of the perspectives and needs of those who use sign languages daily. We recognize that developing such systems without meaningful engagement from the Deaf and Hard-of-Hearing communities risks producing technology that does not serve their interests. Throughout the development of this work, we have collaborated with members of the signing community to ensure our approach remains grounded in authentic linguistic practices.

We use three datasets: ASL STEM Wiki \citep{yin2024asl}, a licensed dataset, and Aesop's Fables \citep{aesop1912gutenberg}. These datasets consist of educational, scientific, and traditional narrative content, and do not contain personally identifiable information or offensive material.

\bibliography{custom}

\appendix

% !TEX root = ../main.tex

\appendix

\section{Translation Prompts}
\label{app:prompts}

This appendix presents the LLM prompts used in each experimental condition.
Unless stated otherwise, all experiments use Gemini 2.5 Pro as the underlying model; the backbone comparison in \S\ref{subsec:backbone} uses Qwen3.6-35B-A3B with these same prompts.
All three conditions require the model to return a JSON object containing:
(1)~the ASL gloss translation,
(2)~spatial index--entity mappings with corresponding English words,
(3)~directional verb source/target mappings,
(4)~whether a rhetorical question (QAC) was used, and
(5)~concept-to-gloss mappings for nouns.
The output format specification is shared across conditions and is listed separately in Section~\ref{app:output-format}.

A vocabulary constraint is enforced in all conditions: every gloss must come from a provided vocabulary list, with fingerspelling (\texttt{fs-WORD}) as the only fallback.
For all conditions, sentences within a story are processed sequentially (causally).
The system and user message portions of the prompt are concatenated into a single prompt string.

\subsection{Sentence-Level Baseline}
\label{app:prompt-sentence-level}

The sentence-level baseline translates each sentence independently with no prior context and minimal instructions.

\medskip
\noindent\hrulefill

\smallskip
\noindent
You are an ASL translator. Translate English sentences to ASL gloss format.

\smallskip
\noindent
VOCABULARY CONSTRAINT: Use only glosses from this vocabulary: \textit{\{vocabulary\}}

\smallskip
\noindent
ASL GRAMMAR BASICS:

\noindent
-- Use spatial indices for entities (IX-3p:i, IX-3p:j, IX-loc:k)

\noindent
-- Use directional verbs when appropriate (i:GIVE:j, 1p-MEET:i)

\noindent
-- You may use rhetorical questions (QAC) if the sentence structure suggests it

\smallskip
\noindent
Translate: ``\textit{\{current\_sentence\}}''

\smallskip
\noindent\hrulefill

\subsection{Context-Aware Baseline}
\label{app:prompt-context-aware}

\smallskip
The context-aware baseline uses the same minimal instructions as the sentence-only baseline but adds a listing of all previously translated sentences and their ASL translations, without explicit guidance on how to use this context.

\medskip
\noindent\hrulefill

\smallskip
\noindent
You are an ASL translator. Translate English sentences to ASL gloss format using discourse context.

\smallskip
\noindent
VOCABULARY CONSTRAINT: Use only glosses from this vocabulary: \textit{\{vocabulary\}}

\smallskip
\noindent
ASL GRAMMAR BASICS:

\noindent
-- Use spatial indices for entities (IX-3p:i, IX-3p:j, IX-loc:k)

\noindent
-- Use directional verbs when appropriate (i:GIVE:j, 1p-MEET:i)

\noindent
-- You may use rhetorical questions (QAC) if appropriate

\smallskip
\noindent
PREVIOUS CONTEXT:

\noindent
Sentence 1: ``\textit{\{sentence$_1$\}}'' $\rightarrow$ \textit{\{translation$_1$\}}

\noindent
Sentence 2: ``\textit{\{sentence$_2$\}}'' $\rightarrow$ \textit{\{translation$_2$\}}

\noindent
\ldots

\smallskip
\noindent
CURRENT SENTENCE: ``\textit{\{current\_sentence\}}''

\smallskip
\noindent\hrulefill

\subsection{Proposed Approach}
\label{app:prompt-proposed}

\smallskip
The proposed approach uses a detailed prompt with explicit discourse-level translation principles and module-specific state registries.
It is composed of two parts: a system message with translation guidelines, and a per-sentence user message that includes (1)~a configurable context window of $w$ previous sentence--translation pairs, and (2)~structured registries corresponding to the three modules described in \S\ref{subsec:framework_architecture}. The registries are accumulated from the LLM's own structured output across previous sentences and passed back as explicit constraints. Below we present the combined prompt with redundancies between the two parts removed for brevity.

\medskip
\noindent\hrulefill

\smallskip
\noindent
You are an expert ASL (American Sign Language) translator specializing in discourse-level translation.
You process sentences sequentially, maintaining consistency with previous translations while having no
knowledge of future sentences.

\smallskip
\noindent
CRITICAL VOCABULARY CONSTRAINT:
Every gloss in your output must exist in the provided vocabulary. This is a hard requirement---do not use
any glosses outside the vocabulary list. If no suitable vocabulary gloss exists for a concept, use
fingerspelling with the format (fs-WORD) as the only fallback.

\smallskip
\noindent
Available vocabulary: \textit{\{vocabulary\}}

\smallskip
\noindent
CORE ASL GRAMMAR PRINCIPLES:

\noindent
1. Use Topic-Comment structure when appropriate

\noindent
2. Omit copula verbs (``is'', ``are'', ``was'', ``were'')

\noindent
3. Use ASL word order (often SOV or Topic-Comment)

\noindent
4. Maintain spatial consistency for referents throughout discourse

\noindent
5. Use appropriate non-manual markers and classifiers

\smallskip
\noindent
SPATIAL INDEXING SYSTEM:

\noindent
-- First person: IX-1p (I/me), POSS-1p (my/mine)

\noindent
-- Second person: IX-2p (you), POSS-2p (your), IX-2p-pl-arc (you plural)

\noindent
-- Third person: IX-3p:i, IX-3p:j, IX-3p:k (different spatial locations)

\noindent
-- Locations: IX-loc:i, IX-loc:j, IX-loc:k (specific places)

\noindent
-- Directional verbs: 1p-GIVE:i, i:GIVE:j, \\ i:MEET:j, etc.

\smallskip
\noindent
For each spatial index, specify which entity it represents and which English word/phrase in the current sentence it corresponds to. If carried over from previous context, use ``implicit''.

\smallskip
\noindent
RHETORICAL QUESTIONS (QAC) USAGE:

\noindent
Use rhetorical questions for:

\noindent
1. Cause-Effect: Replace ``because'', ``so that'', ``in order to''

\noindent
\quad Example: ``I study because I want good grades'' $\rightarrow$
``IX-1p STUDY WHY? GRADE GOOD WANT''

\noindent
2. Topic Introduction: Introduce new information or shift focus

\noindent
3. Emphasis: Highlight important information

\smallskip
\noindent
Do not use rhetorical questions at discourse beginning, when directly answering a genuine question, or for simple statements that don't need emphasis.

\smallskip
\noindent
CONCEPT-GLOSS MAPPING:
For each sentence, identify all nouns in the English sentence and
specify which ASL gloss you used to translate each noun. Only include nouns that appear in your translation.

\smallskip
\noindent
PREVIOUS DISCOURSE CONTEXT:

\noindent
Sentence 1: ``\textit{\{sentence$_1$\}}''

\noindent
ASL Translation 1: \textit{\{translation$_1$\}}

\smallskip
\noindent
Sentence 2: ``\textit{\{sentence$_2$\}}''

\noindent
ASL Translation 2: \textit{\{translation$_2$\}}

\smallskip
\noindent
\ldots

\smallskip
\noindent
SCM --- SPATIAL INDEX REGISTRY $\mathcal{S}$ (established assignments):

\noindent
Spatial Consistency constraint: You MUST use the same spatial index for the same entity.

\noindent
Coreference Correspondence constraint: Each spatial reference must correspond to an entity mention in the English source.

\smallskip
\noindent
Established entity-index assignments:

\noindent
-- IX-3p:i $\rightarrow$ fox

\noindent
-- IX-3p:j $\rightarrow$ rabbit

\noindent
-- IX-loc:k $\rightarrow$ forest

\noindent
\ldots

\noindent
Assign new spatial indices only for entities not yet in this registry.

\smallskip
\noindent
Established directional verb patterns:

\noindent
-- i:GIVE:j: source=fox, target=rabbit

\noindent
\ldots

\noindent
Both source and target indices in directional verbs must satisfy the spatial consistency constraint.

\smallskip
\noindent
QACM --- QAC DECISION REGISTRY $\mathcal{Q}$ (prior QAC usage):

\noindent
Discourse Position constraint: QACs should NOT appear at discourse beginning or as responses to genuine questions.

\noindent
Contextual Appropriateness constraint: Use QACs only when the information structure warrants topicalization (causal explanations, contrastive elements, emphasis).

\smallskip
\noindent
Prior QAC decisions:

\noindent
-- Sentence 1: no QAC

\noindent
-- Sentence 2: QAC used

\noindent
\ldots

\smallskip
\noindent
CGCM --- CONCEPT-GLOSS REGISTRY $\mathcal{L}$ (established mappings):

\noindent
Mapping Consistency constraint: You MUST use the established gloss for any concept already in this registry.

\noindent
Vocabulary Compliance constraint: All glosses must come from the provided vocabulary.

\smallskip
\noindent
Established concept-gloss mappings:

\noindent
-- ``\textit{fox}'' $\rightarrow$ \textit{FOX} (established in sentence 1)

\noindent
-- ``\textit{rabbit}'' $\rightarrow$ \textit{RABBIT} (established in sentence 1)

\noindent
-- ``\textit{forest}'' $\rightarrow$ \textit{fs-F-O-R-E-S-T} (established in sentence 1)

\noindent
\ldots

\noindent
For NEW concepts not in this registry, select an appropriate gloss from the vocabulary. It will be registered for future sentences.

\smallskip
\noindent
CURRENT SENTENCE TO TRANSLATE: ``\textit{\{current\_sentence\}}''

\smallskip
\noindent\hrulefill

\medskip
\noindent
The three registries are accumulated from the LLM's own structured output across previous sentences within the context window: \texttt{spatial\_mappings} and \texttt{directional\_verbs} populate $\mathcal{S}$, \texttt{qac\_used} populates $\mathcal{Q}$, and \texttt{concept\_mappings} populates $\mathcal{L}$. For concept mappings, the first occurrence of each concept within the window establishes the canonical gloss; for spatial indices, the most recent assignment per index is used. This ensures that each module's constraints (\S\ref{subsubsec:scm}--\ref{subsubsec:cgcm}) are implemented as explicit prompt-level constraints rather than relying on the LLM to infer consistency from raw translation context.

\smallskip
\noindent
The context window parameter $w$ controls how many previous sentence--translation pairs are included, and the module registries are bounded by the same window.
When $w = 0$, no context or registry state is provided; when $w = n$, the registries reflect the $n$ most recent sentences; when $w$ is unset, all previous sentences contribute to both context and registries.

\paragraph{Post-Processing Verification}
\label{app:verification}

In addition to the prompt-level registry constraints, a programmatic verification step runs after each LLM call to enforce module constraints as hard constraints. The verification checks each module's output against the accumulated registries and applies corrections where violations are detected:

\begin{itemize}[noitemsep]
    \item \textit{SCM}: If an entity's spatial index differs from its registry assignment, the index is corrected via string replacement throughout the translation, and all affected spatial mappings and directional verbs are updated accordingly.

    \item \textit{CGCM}: If a concept's gloss differs from the registry's established mapping, the gloss is replaced throughout the translation and the concept mappings are updated.

    \item \textit{QACM}: The discourse position constraint ($t{=}0$) is enforced via pre-processing: the prompt for the first sentence explicitly prohibits QAC usage. As a fallback, if the LLM still reports a QAC at $t{=}0$, the metadata is corrected.
\end{itemize}

\noindent
All corrections are applied deterministically without re-querying the LLM. The SCM and CGCM operate via post-processing string replacement; the QACM operates via pre-processing prompt construction based on sentence position.

\subsection{Output Format}
\label{app:output-format}

All three conditions include the following output format specification in their prompts.

\medskip
\noindent\hrulefill

\smallskip
\noindent
Return a JSON with the following fields:

\smallskip
\noindent
-- ``translation'': the ASL gloss translation

\noindent
-- ``spatial\_mappings'': a mapping from each spatial index to its entity name and corresponding English word, e.g., \{``IX-3p:i'': \{``entity'': ``fox'', ``english\_word'': ``he''\}\}

\noindent
-- ``directional\_verbs'': a mapping from each directional verb to its source and target entities with corresponding English words, e.g., \{``i:GIVE:j'': \{``source'': ``fox'', ``target'': ``rabbit'', ``source\_word'': ``he'', ``target\_word'': ``her''\}\}

\noindent
-- ``qac\_used'': boolean indicating whether a rhetorical question structure was used

\noindent
-- ``concept\_mappings'': a mapping from English nouns to their ASL glosses, e.g., \{``vehicle'': ``CAR''\}

\smallskip
\noindent\hrulefill

\subsection{Back-Translation Prompt}
\label{app:prompt-back-translation}

Back-translation is used to produce English reconstructions from ASL gloss output for computing chrF and COMET scores. Sentences are translated causally: each sentence is back-translated with access to all previous ASL sentences in the story but no future sentences, mirroring the forward translation setup.

\medskip
\noindent\hrulefill

\smallskip
\noindent
You are an expert ASL interpreter with deep knowledge of ASL discourse structure, spatial referencing, and narrative coherence. You are translating ASL gloss sequences back to natural English while maintaining discourse consistency.

\smallskip
\noindent
TASK: Translate the TARGET ASL sentence back to natural English, using the previous story context to ensure proper discourse coherence, pronoun resolution, and narrative flow.

\smallskip
\noindent
Previous ASL sentences in this story:

\noindent
1. \textit{\{translation$_1$\}}

\noindent
2. \textit{\{translation$_2$\}}

\noindent
\ldots

\smallskip
\noindent
Now translate the next sentence:

\smallskip
\noindent
TARGET ASL SENTENCE TO TRANSLATE:

\noindent
\textit{\{sentence\_index\}}. \textit{\{target\_asl\}}

\smallskip
\noindent
TRANSLATION GUIDELINES:

\noindent
1. Use the previous story context to resolve spatial references (IX-3p:i, IX-3p:j, etc.) to appropriate English pronouns or noun phrases

\noindent
2. Maintain narrative coherence with previous sentences

\noindent
3. Translate to natural, fluent English that continues the story flow

\noindent
4. Preserve the meaning and discourse function of the ASL sentence

\noindent
5. Use appropriate English discourse markers and connectives when needed

\noindent
6. Ensure pronoun antecedents are clear from the established context

\noindent
7. Maintain consistency with spatial reference assignments from previous sentences

\smallskip
\noindent
Return ONLY a JSON: \{``translation'': ``your\_english\_translation''\}

\smallskip
\noindent\hrulefill

\subsection{Coreference Chain Extraction Prompt}
\label{app:prompt-coreference}

English coreference chains are extracted from source stories to serve as the ground-truth entity structure for computing SCA. Unlike the causal translation and back-translation prompts, this prompt receives the entire story at once, as coreference resolution requires full document context.

\medskip
\noindent\hrulefill

\smallskip
\noindent
You are an expert in coreference resolution. Analyze the text for coreference chains and provide detailed output.

\smallskip
\noindent
Pay special attention to:

\noindent
-- First-person pronouns (I, me, my, mine) and their referents

\noindent
-- Narrator identity in first-person narratives

\noindent
-- All pronouns and their antecedents

\noindent
-- Ambiguous pronouns like ``it'' that might refer to different entities

\smallskip
\noindent
Sentences:

\noindent
1. \textit{\{sentence$_1$\}}

\noindent
2. \textit{\{sentence$_2$\}}

\noindent
\ldots

\smallskip
\noindent
Return your analysis as JSON:

\smallskip
\noindent
\{``chains'': [\{``chain\_id'': 1, ``canonical\_entity'': ``first\_mention'',

\noindent
\quad ``mentions'': [``mention1'', ``mention2'', \ldots],

\noindent
\quad ``sentence\_locations'': [1, 2, \ldots]\}]\}

\smallskip
\noindent\hrulefill

\section{Human Evaluation Details}
\label{app:human_eval}

\paragraph{Study Design} We selected 32 stories (152 sentences) from Aesop's Fables via stratified sampling across automated SCA, CGC, and QAC score ranges (low: $<$0.33, medium: 0.33--0.67, high:
$\geq$0.67), drawing equally from the proposed framework and sentence-level baseline outputs. Two ASL-fluent evaluators independently rated each sentence on a 1--5 Likert scale for spatial
coreference accuracy (SCA), concept-gloss consistency (CGC), and QAC appropriateness (QAC$_{\text{Ap}}$), with an N/A option when QAC was not applicable.

\paragraph{Inter-Rater Agreement} For SCA, inter-rater agreement is moderate: weighted $\kappa$\,=\,0.43 (quadratic), Spearman $\rho$\,=\,0.49 ($p$\,$<$\,0.001), with 64\% within-one-point agreement across 144 paired ratings. For CGC and QAC, both evaluators assigned near-ceiling ratings (CGC means: 4.72 and 4.96; QAC means: 4.58 and 5.00), resulting in insufficient variance for meaningful agreement
statistics.

\paragraph{Correlation Analysis} We report story-level Spearman correlations (mean aggregation across sentences), as sentences within a story are not independent. Table~\ref{tab:human_eval_full} presents per-evaluator and combined results.

\begin{table}[h]
\centering
\small
\begin{tabular}{llcc}
\textbf{Metric} & \textbf{Evaluator} & $\rho$ & $p$ \\
\midrule
\multirow{3}{*}{SCA} & Evaluator 1 & 0.50 & 0.005 \\
& Evaluator 2 & 0.42 & 0.022 \\
& Combined & 0.52 & 0.003 \\
\midrule
\multirow{3}{*}{CGC} & Evaluator 1 & 0.48 & 0.007 \\
& Evaluator 2 & 0.24 & 0.195 \\
& Combined & 0.48 & 0.007 \\
\midrule
\multirow{3}{*}{QAC$_{\text{Ap}}$} & Evaluator 1 & $-$0.12 & 0.561 \\
& Evaluator 2 & 0.28 & 0.301 \\
& Combined & $-$0.09 & 0.659 \\
\bottomrule
\end{tabular}
\caption{Story-level Spearman correlations ($\rho$) between automated metrics and human ratings. Combined averages both evaluators' ratings per sentence before correlating.}
\label{tab:human_eval_full}
\end{table}

\paragraph{QAC Subjectivity} QAC$_{\text{Ap}}$ correlations are non-significant for both evaluators. We attribute this primarily to the inherently subjective nature of QAC appropriateness judgments. Our automated
metric evaluates QAC usage against specific discourse criteria: the presence of causal relationships, topicalization, or emphasis markers in the English source. However, native ASL signers may find
QAC restructuring acceptable even in the absence of such markers. For example, the English sentence ``There are two sides to every question'' can be rendered as either \textit{EACH
QUESTION HAVE TWO SIDE} or \textit{EACH QUESTION HAVE WHAT? TWO SIDE}. The latter employs a QAC for topicalization that is natural in ASL discourse but has no explicit discourse trigger in the
English source. This gap between the metric's rule-based criteria and the broader range of acceptable human judgments makes strong automated-human agreement challenging for this phenomenon. The evaluators noted that some QAC usages were acceptable even when not matching their personal signing preferences. We further examined agreement on QAC \textit{usage} (binary presence/absence, as opposed to appropriateness). At the story level, one evaluator shows strong agreement with the system's self-reported
  \textit{qac\_used} field ($\rho$\,=\,0.85, $p$\,$<$\,0.001), while the other shows moderate agreement ($\rho$\,=\,0.39, $p$\,=\,0.035). Inter-rater agreement on QAC presence is similarly moderate
  ($\rho$\,=\,0.41, $p$\,=\,0.001). The primary source of disagreement is that one evaluator rated 28 sentences as not containing a QAC (N/A) where both the system and the other evaluator identified
  one, suggesting differing thresholds for what constitutes a QAC structure.

\noindent
One direction for future work is to replace the QACM's deterministic trigger rules with a learned trigger model, trained on annotated data where QAC structures are identifiable from gloss patterns (a \textit{wh}-constituent followed by an answer clause). Emitting a confidence score rather than a binary decision would allow restructuring to be applied only in high-confidence contexts, with mid-range cases treated as stylistic variants. A further extension would condition on signer preference and treat QAC frequency as a register parameter, analogous to formality in text generation.

\paragraph{CGC Ceiling Effects} CGC shows significant combined correlation ($\rho$\,=\,0.48, $p$\,=\,0.007), driven primarily by Evaluator 1 ($\rho$\,=\,0.48, $p$\,=\,0.007). Evaluator 2 shows a positive but non-significant trend ($\rho$\,=\,0.24, $p$\,=\,0.195), likely attenuated by near-ceiling ratings (95\% rated 5). Both evaluators assigned high CGC ratings overall (94--95\% rated 4 or 5), which limits discriminative power. We note that CGC evaluates cross-sentence gloss consistency (whether the same concept receives the same gloss throughout the discourse), which is a subtler judgment than per-sentence
translation accuracy and may be difficult to assess reliably without explicit side-by-side comparison tools.

\noindent
The ceiling effect appears on the human-rating side rather than in the metric itself. CGC remains discriminative on the translation side: it is 0.59 for the sentence-level baseline (Table~\ref{tab:discourse_results}) and moves from 0.64 to 0.97 as the CGCM is enabled (Table~\ref{tab:system_ablation}). Validating CGC more sharply would require sampling translations with a wider spread of consistency violations than the systems we evaluate produce.

\end{document}